\documentclass{article}
        
\usepackage[nonatbib, final]{nips_2017}

\usepackage[backend=biber,
            hyperref=true,
            url=false,
            isbn=false,
            doi=false,
            backref=true,
            style=numeric-comp,
            maxbibnames=50,
            mincitenames=1,
            maxcitenames=2,
            sorting=nyt,
            block=none]{biblatex}

\usepackage{graphics}
\usepackage[pdftex]{graphicx}
\usepackage{wrapfig}
\DeclareGraphicsExtensions{.pdf,.png,.jpg}

\usepackage[font=small]{caption}
\usepackage{subcaption}
\usepackage[rightcaption]{sidecap}
\usepackage{pbox}

\usepackage{mathtools}
\usepackage{amsmath, amssymb, amscd}
\usepackage{ wasysym } 
\usepackage{amsfonts}       
\usepackage{mathptmx} 
\usepackage{gensymb} 
\usepackage{nicefrac}       

\DeclareMathAlphabet{\mathcal}{OMS}{lmsy}{m}{n}
\DeclareSymbolFont{largesymbols}{OMX}{cmex}{m}{n}
\usepackage{textcomp} 

\usepackage{algorithm} 

\usepackage{algorithmic} 
\floatname{algorithm}{Algorithm}

\usepackage{booktabs}       
\usepackage{array} 
\usepackage{tabularx}
\usepackage{multirow}
\usepackage{multicol}
\usepackage{booktabs}

\usepackage[utf8]{inputenc}
\usepackage[T1]{fontenc} 
\usepackage[english]{babel} 
\usepackage{units}
\usepackage{bm}
\usepackage{times} 
\usepackage{xspace}
\usepackage{flushend}
\usepackage{csquotes}
\usepackage{makeidx}
\usepackage{enumitem}
\setlist[enumerate]{itemsep=0mm}
\setitemize[0]{leftmargin=10pt}
\usepackage{soul} 
\usepackage{subfiles} 

\usepackage[protrusion=true,expansion=true]{microtype}
\usepackage[yyyymmdd]{datetime}

\date{\protect\formatdate{1}{1}{2001}}

\usepackage{url}
\usepackage{hyperref}
\hypersetup{
    colorlinks=true,
    linkcolor=black,
    citecolor=black,
    filecolor=cyan,
    urlcolor=black
}
\makeatletter
\g@addto@macro{\UrlBreaks}{\UrlOrds}
\makeatother
\usepackage{color}
\usepackage[usenames,dvipsnames,table]{xcolor}

\usepackage{soul} 

\usepackage{xspace} 
\newcommand{\etal}{\textit{et al}.\xspace}

\newcommand{\ignore}[1]{}

\newcommand{\figref}[1]{Figure~\ref{fig:#1}}

\newcommand{\seclabel}[1]{\label{sec:#1}}

\newenvironment{tight_itemize}{
\begin{itemize}[leftmargin=10pt]
  \setlength{\topsep}{0pt}
  \setlength{\itemsep}{2pt}
  \setlength{\parskip}{0pt}
  \setlength{\parsep}{0pt}
}{\end{itemize}}

\title{DeformNet: Free-Form Deformation Network\\ for 3D Shape Reconstruction from a Single Image}

\author{
  Andrey Kuryenkov\thanks{Equal Contribution}, Jingwei Ji$^*$, Animesh Garg, Viraj Mehta, JunYoung Gwak, \\ \textbf{Christopher Choy, Silvio Savarese}\\
  Stanford Vision and Learning Lab
}

\newcommand{\algoName}{\textsc{DeformNet}\xspace}

\begin{document}

\maketitle

\begin{abstract} 
3D reconstruction from a single image is a key problem in multiple applications ranging from robotic manipulation to augmented reality. Prior methods have tackled this problem through generative models which predict 3D reconstructions as voxels or point clouds. However, these methods can be computationally expensive and miss fine details. We introduce a new differentiable layer for 3D data deformation and use it in \algoName to learn a model for 3D reconstruction-through-deformation. \algoName takes an image input, searches the nearest shape template from a database, and deforms the template to match the query image. We evaluate our approach on the ShapeNet dataset and show that - (a) the Free-Form Deformation layer is a powerful new building block for Deep Learning models that manipulate 3D data (b) \algoName uses this FFD layer combined with shape retrieval for smooth and detail-preserving 3D reconstruction of qualitatively plausible point clouds with respect to a single query image (c) compared to other state-of-the-art 3D reconstruction methods, \algoName quantitatively matches or outperforms their benchmarks by significant margins. For more information, visit: \href{https://deformnet-site.github.io/DeformNet-website/}{\textcolor{blue}{https://deformnet-site.github.io/DeformNet-website/}.
}

\end{abstract}

\vspace{-10pt}
\section{Introduction}
\seclabel{intro}
\vspace{-5pt}
This paper studies the structured prediction problem of regressing unordered point sets with implicit and often ambiguous input spaces. A concrete instance which embodies this type of problem is 3D object geometry reconstruction (3DR) from single-view images for partial shape guidance \cite{firman2016structured}. 
The ambiguity arises from the fact that 3D-to-2D mapping is not invertible and large portions of the object features are typically occluded. 
3DR is a pivotal learning problem in visual understanding, with numerous applications across domains. For instance, an intelligent robot requires a 3D model of the object instance to reason about manipulation. Similarly, in augmented reality recognizing the 3D shapes of often unseen objects in the world is necessary for both correct rendering and interaction. 

3DR has been explored in a large body of extant work in computer vision, for problems such as structure from motion \cite{sfm,slam} or multiview stereo \cite{Furukawa_etal_2010,Furukawa_Ponce_2010,Goesele_etal_2010,Hernandez_Vogiatzis_2010, lhuillier2005quasi, agarwal2009building, engel2014lsd} and at times even with single view images~\cite{Criminisi00a}.
Ingenious work on ``Shape-from-X" has utilized priors on natural images to infer geometric features, with ``X" being shading, texture, specularity, shadow and so on \cite{zhang1999shape, malik1997computing, healey1988local, savarese20073d, BarronTPAMI2015}. Most of the aforementioned methods require carefully constructed features, a problem that is addressed by data-driven methods relying on large-scale 3D object datasets \cite{DBLP:journals/corr/ChangFGHHLSSSSX15, xiang2016objectnet3d}.

Data-driven methods learn implicit priors for various object recognition tasks such as shape completion and 3D reconstruction. Broadly, these methods use the prior knowledge in two ways: (i) image-based shape retrieval that focuses on algorithm design to find the nearest shape in database for the query image \cite{xiang2014beyond, DBLP:journals/corr/ChangFGHHLSSSSX15,rock2015completing}, and (ii) deep generative models which operate directly on the query image and generate a 3D reconstruction as output, matching the shape distribution but resulting in different shape instances than in the database \cite{choy20163d, fan2016point, rezende2016unsupervised, wu2016single, yan2016perspective,drcTulsiani17,wsgan}. 

We note that the majority of recent methods for 3DR resort to either direct volumetric representation (aka voxels) or meshes from multi-view images as their shape representation. While intuitive, these representations can be both computationally inefficient and ineffective in capturing the natural invariance of 3D shapes under geometric transformations. A recent method by Fan et al. uses Point Set Generation Network (PSGN) to alleviate these problems. As they note, a point cloud is a simple, uniform structure that is relatively easier to learn than voxels, as it does not have to encode multiple primitives or combinatorial connectivity patterns. Additionally, point clouds are computationally superior to voxels since they do not require 3D convolutions and are amenable to direct manipulation when it comes to shape deformation and transformation. However, though direct prediction of fixed size point clouds improves 3DR performance, giving up on connectivity can result in a lack of fine shape features since loss functions are focused on overall reconstruction. 

At the same time, Spatial Transformer Networks (STN) have presented an appealing method to learn geometric transformations in 2D images~\cite{jadenberg2015stn}. STNs are a modular, differentiable and dynamic upgrade to pooling operations that learn to zoom in and eliminate background clutter, thereby “standardizing” the input. However, they have primarily been studied in the context of discrete grids in image inputs to facilitate image classification. 

\begin{figure}[t]
\centering
 \includegraphics[width=0.9\linewidth, trim={0.5cm 11.5cm 7cm 3.2cm},clip]{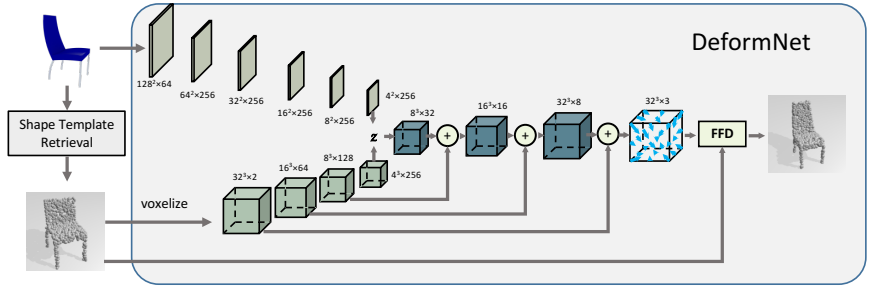}
 \caption{Our framework and DeformNet architecture. `+' denotes stacking activations from image encoder, voxel encoder and voxel decoder. The output of decoder is the prediction on the offsets of control points, which decides the free-form deformation of input shape on the next step.}
 \label{fig:archFig}
 \vspace{-15pt}
\end{figure}

Inspired the ideas from PSG and STN, we propose \algoName - a model that extends STN style geometric operations to 3D using the notion of Free-Form Deformations (FFD). When used in conjunction with a point cloud representation, this method not only benefits from computational efficiency but also can preserve fine details in shapes since it implicitly preserves connectivity in structures.
\algoName uses a single image input to first perform shape retrieval from an object dataset using a learned image-to-shape embedding, and then deforms the point cloud representation of the retrieved template using the FFD layer in an encoder-decoder style network architecture as depicted in \figref{archFig}.
\algoName intuitively builds on the strengths of Shape Retrieval, PSG, and STNs while compensating for their shortcomings. We implement FFD as a differentiable layer for end-to-end training along with point set correspondence based loss functions Chamfer Distance and Earth-Mover's Distance, as in \cite{fan2016point}. 

To summarize, the main contributions of this paper are:
\vspace{-5pt}
\begin{tight_itemize}
\item Introduction of Free Form Deformation as a differentiable layer to be used as a new building block that enables 3D data manipulation.
\item A novel framework, \algoName based on FFD layer to achieve smooth geometric deformations in point clouds for 3D Reconstruction.
\item Evaluation of \algoName on rendered images achieves state-of-the-art performance in comparison to both Point Prediction Methods such as PSG~\cite{fan2016point} and Generative Models such as 3D-R2N2~\cite{choy20163d}.
\end{tight_itemize}

\vspace{-5pt}
\section{Related Work}
\vspace{-5pt}

\noindent \textbf{Generative Models for 3D Reconstruction.} Recently,  generative models for 3D reconstruction have produced state-of-the-art results. One approach is targeting voxel reconstruction through a 3D voxel neural network. \cite{choy20163d} proposed a 3D recurrent neural network (3D-R2N2) based on long-short-term memory (LSTM) to infer 3D volumetric shape from a single view or multiple views. Girdhar \etal \cite{tlnet} proposed a TL-embedding network which embeds image and shape together for single view 3D volumetric shape generation. Wu \etal \cite{3dgan} proposed a 3D VAE-GAN which brings the two popular generative models together in volumetric shape generation and reconstruction. There are also representative advances in unsupervised/weakly supervised 3D learning for reconstruction. \cite{yan2016perspective} proposed a perspective transformer net for reconstruction from a single image which only uses images contour as supervision. \cite{rezende2016unsupervised} proposed a conditional generative network for unsupervised reconstruction from images.

All of the above neural network based 3D reconstruction methods use voxel representations, which requires a large amount of memory and is inefficient due to the small space usage of generic 3D shapes. Fan \etal \cite{fan2016point} proposed an alternative approach with neural networks that output unordered 3D point sets for 3D reconstruction. In our work, we combine both voxel and point set representations by using a 3D convolutional neural network to generate a deformation that is applied to 3D points sets for output which can preserve fine detail in the template shape.

\noindent \textbf{Spatial Transformer Networks} Conventional convolutional neural networks lack the ability to warp or select a patch from an image that is relevant to the target task, which leads to an unnecessarily larger and deeper network for larger images. To alleviate this issue, Jadenberg \etal~\cite{jadenberg2015stn} proposed the Spatial Transformer Networks (STN) that can apply geometric transformations to an input or activations such as Affine Transformation or Thin-Plane-Spline to extract a patch that is relevant to the task. Kanazawa \etal \cite{warpnet} proposed a WarpNet, an unsupervised method for deforming an image using a Spatial Transformer Network. This work is similar in spirit to the WarpNet in that the neural network generates deformation parameters and the loss is defined using the deviation of the deformed input to the ground truth. However, unlike the WarpNet, we condition the network on two inputs (the reference image and a template shape) and use it for 3D shape deformation and reconstruction. 

\noindent \textbf{3D Shape Deformation}
Recently, Yumer and Mitra \cite{yumer2016learning} proposed a 3D Convolutional Neural Network that generates a deformation field as an output for 3D mesh editing given a user input. We employ the same deformation representation, Free-Form Deformation (FFD) to generate 3D deformation field. 

The primary difference between our approach and that of Yumer \etal are that they supervise the network using precomputed deformation offsets while we propose an end-to-end trainable network which only requires the target shape as the sole source of supervision. We accomplish this by computing distance between a set of points and minimizing it with respect to the deformation field (Sec.~\ref{sec:loss}). In addition, unlike Yumer \etal, we focus on the 3D reconstruction given an image input, rather than 3D editing.

Huang \etal \cite{huang2015single} also proposed an approach to 3D reconstruction through template retrieval and deformation, which relies on jointly segmenting the 2D image and 3D templates and creating the 3D reconstruction from deformed parts of the segmented templates. Their approach differs from ours in that it relies on having noisy segmentation of the shapes in their database, in that it performs reconstruction by segmenting both the input image and templates and deforming template parts to fit with the parts of the segmented image via direct optimization, and in that it takes significantly longer to run due to the optimizations it employs. 

To summarize, voxelized representations suffer from memory inefficiency and difficulty in generating fine details, previous deformation networks require the limiting supervision of point displacements, and previous work on reconstruction through template deformation relied on noisy segmentation in the shapes database. Purely generative models don't leverage the high quality and broad availability of shape databases. Our paper finds a balance between the flexibility of a generative approach and the output quality of a database-focused approach. It does this without requiring any supervision beyond that of the desired shape. This addresses the major problems with the current best methods.

\vspace{-15pt}
\section{\algoName}
\vspace{-5pt}
In this section, we propose the \algoName framework that generates a 3D shape reconstruction from a single image. Unlike recent single-view 3D reconstruction works that use Convolutional Neural Networks, we do not use voxelized output \cite{choy20163d,tlnet,3dgan}, which cannot recover fine details due to the coarse resolution of the voxel grid, and do not directly generate a point cloud from scratch \cite{fan2016point}. Instead, we propose an end-to-end network that deforms a template 3D shape to match an input image and preserves the topology of the template shape and train it using the target shape only. This can be seen as combining the key insights from prior work that used Free-Form Deformation in a deep learning model \cite{yumer2016learning} and the Chamfer distance for the objective function \cite{fan2016point}.

In the following subsections, we first introduce metric learning based shape template retrieval, which finds the most topologically similar 3D shape template for a given image (Sec.~\ref{sec:shape-retrieval}).
Then, we introduce {\algoName}, which takes a set of templates that we use for deformation and a query image and learns to output the deformation field end-to-end given the target shape (Sec.~\ref{sec:ffd}). Lastly, we present two objective functions that measure the deviation of the deformed shape from the target shape by finding correspondences on-the-fly (Sec.~\ref{sec:loss}). See Figure \ref{fig:archFig} for illustration.

\vspace{-5pt}
\subsection{Shape Template Retrieval}
\label{sec:shape-retrieval}

To make use of the high-quality 3D CAD models in the existing database, the first step of the framework is to retrieve shape templates that have a similar topology to the object in a query image. For this, we use metric learning to learn an embedding that preserves topological similarity between shapes.
Specifically, we first define the metric space by a set of constraints that force dissimilar shapes to be at least a margin father away than the distance between similar objects.
\begin{equation}
d(F(x_i; \theta), F(x_j; \theta)) + \Delta < d(F(x_i; \theta), F(x_k;\theta)) \text{ where } i, j \in \mathcal{C}_n, k \in \mathcal{C}_m, m \neq n
\end{equation}
where $\Delta \in \mathbb{R}$ is a margin, $\mathcal{C}_n$ indicates a set containing all elements in the $n$-th class and $d(\cdot, \cdot)$ is an arbitrary distance function. We use a parametric metric space representation using a neural network $F(\cdot; \theta)$ where $\theta$ denote the parameters in the neural network. The above constraints can be converted to a loss function which forms a triplet loss \cite{schroff2015facenet}. By minimizing the loss function with respect to $\theta$, we can generate a feature representation that preserves perceptual similarity in a metric space where the distance operation is meaningful. However, due to difficulty in mining hard negatives and inefficiency of not reusing features in a batch, the naive metric learning approach only yields a moderate result \cite{schroff2015facenet, oh2016deep}.
Instead, we use the smoothed version of the triplet loss that reuses all features in a batch for efficient hard negative mining \cite{oh2016deep} and allows fast and effective training.
\begin{equation}
J = 
\frac{1}{2 | \mathcal{P} |} \sum_{(i, j) \in \mathcal{P}} \left[ \log \left(
\sum_{i, k \in \mathcal{N}_i} \exp\{ \Delta - d_{i, k}\} +
\sum_{j, l \in \mathcal{N}_j} \exp\{ \Delta - d_{j, l}\}
\right) + d_{i, j}
 \right]_+^2
\end{equation}
where $d_{i, j}$ indicates $d(F(x_i; \theta), F(x_j; \theta))$ and $\mathcal{N}_i$ denotes a set of shapes that belong to different categories from the category of $i$.

Specifically, we used a 2D CNN to implement $F(x_i; \theta) \in \mathbb{R}^D$ and $x_i$ denotes rendering of the $i$-th shape. We define the positive pairs $\mathcal{P}$ to be renderings of the same shape from different perspectives and negatives $\mathcal{N}$ to be renderings from different shapes. We define the similarity to be the inverse of the distance in the metric space and retrieve $K$-nearest neighbors from a query image and use the shapes that generated the images for the next stage.

\vspace{-5pt}
\subsection{\algoName: Model}
\vspace{-5pt}
Given a reference image and a shape template that closely matches the object in the input image from the Shape Retrieval stage (Sec.~\ref{sec:shape-retrieval}), we want to generate the parameters of a deformation which transforms the shape template into the shape in the reference image.
Unlike conventional CNN for reconstruction, \algoName takes two different modalities as inputs and thus has two CNNs in these respective domains to encode the inputs. First, for the 2D image, we used 2D CNN for an image encoder $E_I(I) \in \mathbb{R}^{F_I}$ and for the 3D shape input, we voxelized sampled points on the surface to generate dense point cloud and voxelized to feed into a 3D CNN encoder $E_s(S) \in \mathbb{R}^{F_S}$.
We simply combined the information from two modes by stacking the final fully connected layer activation to the last 3D CNN activation along the channel so that we preserve the spatial information from the 3D shape template.
The combined information from both encoders contains all information that we need to know from both inputs and thus call the latent variable $z \in \mathbb{R}^{F_I + F_S}$. $z$ is then fed into a 3D decoder $D(z)$, a 3D Deconvolutional Neural Network, to expand the spatial dimension of the output. Since the 3D encoder loses spatial details as we project the activations from 3D convolution layers to a coarser voxel grid, we use a 3D U-Net structure which is an extension of the 2D U-Net proposed in \cite{yumer2016learning} to recover the details in the output. The U-net has an hourglass shape and skip connections between the 3D encoder and 3D decoder that crosses the latent variable (Fig.~\ref{fig:archFig}).

The final output of the decoder is a vector field $V = \{v_i\}_{i = 1, \ldots, N^3}$, $v \in \mathbb{R}^3$ which is used as the offset for the $N^3$ control points in the Free-Form Deformation Layer (Sec.~\ref{sec:ffd}). Each offset $v_i$, represents $x,y,z$ offset of the corresponding control point. These values are used to then compute the deformed point cloud output, which is the final output of the network, and which can then be compared against the ground truth point cloud of the input image shape.

\vspace{-5pt}
\subsection{Free-Form Deformation Layer}
\label{sec:ffd}\vspace{-5pt}

Free-Form Deformation (FFD) \cite{sederberg1986free} is the 3D extension of a Bezier curve form, which has been widely used for shape deformation. Since it is defined on a 3D grid, FFD fits with 3D convolutional neural network and has been used for generating 3D deformation using a neural network before\cite{yumer2016learning}.
The {\algoName} learns the FFD for every input image - shape template pair, and the predicted FFD on shape template will generate the final deformed shape. In this paper, we mainly focus on manipulating and evaluating reconstructions with point clouds, though the learned FFD could also be applied to other formats.

FFD is formally formulated as following. Let $p = (u, v, w)$ be the normalized point coordinate in the grid and $\Delta_{ijk}$ be the 3D deformation offset at the grid control point $p_{ijk} = (i, j, k)$. Then, the point $p$ after deformation is defined as
\begin{equation}
p' = \sum_{i=0}^l \sum_{j=0}^m \sum_{k=0}^n (p_{ijk}+\Delta_{ijk}) B_{l, i}(u) B_{m, j}(v) B_{n, k}(w)
\end{equation}
where $B_{n,m}(x) = \binom{n}{m} (1-x)^{n-m}x^m$ is a binomial function, and $l,m,n$ are sizes of the control point grid.
The interpolation will mix the displacement of all control points around each data point, generating a smooth deformed output.
The 3D decoder outputs deformation field $V$. Note that $p'$ is differentiable with respect to $v_{ijk}$, which guarantees that the backward propagated gradients could flow from the objective function on the top of FFD, making FFD learnable.

\vspace{-5pt}
\subsection{Objective Functions} 
\label{sec:loss}
\vspace{-5pt}

To make the \algoName end-to-end trainable, we need to define a loss function that optimizes the target task: deforming a template shape to match a target shape.
Ideally, the function should measure difference between deformed shape and a template shape and should return 0 if and only if the deformed shape overlaps the template exactly.
However, 3D shapes are defined by vertices and faces whose accurate topological similarity is difficult to measure. So, rather than measuring the topological similarity, we sample points on the surface of a shape and use the set of points (point cloud) as a surrogate for a shape. Point clouds are easy to manipulate due to their simplicity and efficiency, and we follow \cite{fan2016point} by using distance functions on point clouds as the loss function. If the network generates the correct deformation field $V$ that minimizes the point cloud distances, then the deformed point could will match the target point cloud accurately as well. We explore two point cloud distance functions: Earth Mover's distance (EMD)  and Chamfer distance (CD) \cite{rubner2000earth}. Both distance measures are based on point-wise L-2 distance.

\noindent \textbf{Earth Mover's distance:} 
The EMD is defined as the minimum of sum of distances between a point in one set and a point in another set over all possible permutation of correspondences. To find the minimum, the EMD implicitly solves bipartite matching problem. More formally, given two sets of points $S_1, S_2$, the EMD is defined as
\begin{equation}
d_{EMD}(S_1, S_2) = \min_{\phi: S_1 \rightarrow S_2} \sum_{p \in S_1} \| p - \phi(p) \|_2
\end{equation}
where $\phi$ is some bijection from $S_1$ to $S_2$.

\noindent \textbf{Chamfer distance:} CD is computationally easier than EMD, since it uses sub-optimal matching to determine the pair-wise relation. For each point in one point set, CD simply treat the nearest neighbor in the other set as the image of this point. Formally, CD is defined as
\begin{equation}
d_{CD}(S_1, S_2) = \sum_{p_1\in S_1} \min_{p_2 \in S_2} \|p_1 - p_2\|_2^2 + \sum_{p_2 \in S_2} \min_{p_1 \in S_1} \| p_1 - p_2 \|_2^2
\end{equation}

We use two forms of regularization in addition to the distance function, the first being L1 loss over all point cloud offsets to force the network to deform the template as little as possible, and the second being L2 loss over the difference between neighboring control point offsets to promote smooth deformation.

\vspace{-5pt}
\subsection{Implementation Details}
\vspace{-5pt}
For shape retrieval, we use the multiview rendered images of shape from \cite{choy20163d} to train the network and used the GoogLeNet model \cite{szegedy2015going}. At training and testing time, the template shape input is chosen at random from the 5 most similar shapes from the same category. For EMD, we precomputed ground truth correspondences between each input and its 5 most similar shapes using the Hungarian Algorithm; thus it should be noted that the correspondence is based on the undeformed template, which is an approximation of true EMD. We evaluated using true EMD loss and found it to not work better, and as discussed in the appendix focused on using Chamfer distance as the loss for training. For Chamfer distance, we use the output of the network directly to compute the distance.

We used TensorFlow to implement the networks and used Adam optimizer with momentum term of 0.95. The model is trained with an initial learning rate of {5e-4} and goes down to {5e-5} after 20k iterations. After selection based on performance, we choose a batch size of 16 for training. We use leaky ReLU as an activation function. Note that to make use of EMD, we resample the point clouds to normalize the number of points. We train on point clouds with 1024 points. For deformation, we set N=4 as the number of control points in each dimension that points are computed with (so each point in the deformed output point cloud is a function of $N^3=64$ control points). We use $\lambda=0.05$ for regularization on control point offsets.

\vspace{-5pt}
\section{Experimental Evaluation}
\vspace{-5pt}
\subsection{Experimental Setup}
\vspace{-5pt}

\begin{table}[!b]
\vspace{-15pt}
\centering
\caption{Comparison with point set generation network and 3D-R2N2 on point-set based metrics. 'Rtvr' is the loss when just using one of the top 5 templates at random, without deformation. 'No Rtvr' is our model trained without shape retrieval, to deform a random in-category shape. The numbers are the average point-wise distances. (Lower is better)}
\resizebox{0.8\linewidth}{!}{
\begin{tabular}{c|c|c|c|c|c|ccccc}
\hline
\multirow{2}{*}{Category} & \multicolumn{5}{c|}{\cellcolor[HTML]{CBCEFB}CD}      & \multicolumn{5}{c}{\cellcolor[HTML]{FFC72C}EMD}                                                                     \\ \cline{2-9} 
                          & \textbf{Ours} & Rtrv & No Rtrv & PSGN\cite{fan2016point} & 3D-R2N2\cite{choy20163d} & \multicolumn{1}{c|}{\textbf{Ours}} & \multicolumn{1}{c|}{Rtrv} & \multicolumn{1}{c|}{No Rtrv} & \multicolumn{1}{c|}{PSGN\cite{fan2016point}}  & 3D-R2N2\cite{choy20163d} \\ \hline\hline
airplane                  & \bf{0.10} & 0.15 & 0.20 & 0.14 & -       & \multicolumn{1}{c|}{\bf{0.56}} & \multicolumn{1}{c|}{0.64} & \multicolumn{1}{c|}{0.74} & \multicolumn{1}{c|}{1.15} & -        \\
\rowcolor[HTML]{E0E0E0}
bench                     & \bf{0.10} & 0.24 & 0.21 & 0.21 & -       & \multicolumn{1}{c|}{\bf{0.55}} & \multicolumn{1}{c|}{0.64} & \multicolumn{1}{c|}{0.68} & \multicolumn{1}{c|}{0.98} & -       \\
car                       & \bf{0.09} & 0.14 & 0.13 & 0.11 & -       & \multicolumn{1}{c|}{0.52} & \multicolumn{1}{c|}{0.63} & \multicolumn{1}{c|}{0.63} & \multicolumn{1}{c|}{\bf{0.38}} & -       \\
\rowcolor[HTML]{E0E0E0}
chair                     & \bf{0.13} & 0.19 & 0.27 & 0.33 & -       & \multicolumn{1}{c|}{\bf{0.51}} & \multicolumn{1}{c|}{0.62} & \multicolumn{1}{c|}{0.70} & \multicolumn{1}{c|}{0.77} & -       \\
sofa                      & \bf{0.21} & 0.30 & 0.37 & 0.23 & -       & \multicolumn{1}{c|}{0.77} & \multicolumn{1}{c|}{0.83} & \multicolumn{1}{c|}{0.84} & \multicolumn{1}{c|}{\bf{0.60}} & -       \\ \hline
\rowcolor[HTML]{E0E0E0}
mean                      & \bf{0.13} & 0.20 & 0.24 & 0.20 & 0.71    & \multicolumn{1}{c|}{\bf{0.58}} & \multicolumn{1}{c|}{0.67} & \multicolumn{1}{c|}{0.72} & \multicolumn{1}{c|}{0.78} & 1.02    \\ \hline

\end{tabular}
}
\label{table:1}
\vspace{-10pt}
\end{table}

We train and evaluate our models on the ShapeNet database \cite{shapenet}, which contains a large quantity of manually created and cleaned 3D CAD models. Specifically, we select 5 representative categories to study on: chair, car, airplane, bench and sofa, following Gwak \etal~\cite{wsgan}. The images for training and testing are rendered in various angles to provide synthetic training data for the model. In total, 22,324 shape models are covered, where training/testing split is 80\%/20\%. The 3D CAD objects are originally stored as meshes, so we enriched the dataset via resampling the meshes into point clouds and voxelizing them into voxels.

\vspace{-5pt}
\subsection{3D Shape Reconstruction from RGB Images}
\vspace{-5pt}
We compare our work with point set generation network (PSGN) \cite{fan2016point}, which is the state-of-the-art in deep learning based 3D reconstruction from a single image. PSGN chooses point clouds as the 3D representation, which allows manipulation including geometric transformations and deformations. Also, point clouds can in principle contain more information than voxel representations due to the latter's discrete nature, and point clouds are easy to convert into voxels whereas the other way around will be tricky. Therefore we also target point clouds, though the learned free-form deformation can be applied to both voxel, point cloud, and mesh.

On point clouds, PSGN proposed two metrics for training and evaluating the reconstruction - CD and EMD. To have a fair comparison, we use the same point-set based metrics and follow their experiment setups. We train and test with relatively sparse point clouds with 1024 points, though as will be demonstrated in section 4.6 our model can be applied to dense point clouds despite being trained with sparse ones. To have comparable scaling of distances, during evaluation point clouds are bounded in a hemisphere with a radius of unit 1 and are aligned to their ground plane. Unit 1 is defined as 1/10 of the length of the 3D grid as done in PSGN. We train our models on all five categories with only the CD the loss (as it was found that EMD provided no benefit over CD), and provide both CD and EMD metrics on the test set alongside the same metrics for the newest trained model released by PSGN.

Here we report the per category comparison on both CD and EMD metrics in Table \ref{table:1}. In PSGN, they include the mean value of point-set based metrics from 3D-R2N2 \cite{choy20163d}, thus we also list them here.
On CD metrics, we outperform PSGN and 3D-R2N2 on all categories; on EMD, PSGN achieves better values on the car and sofa categories, but our model performs significantly better in the other three categories as well as the average value. This indicates PSGN has good performance on rotund and less detailed objects, while our strength is on objects with more fine-grained details such as chairs and airplanes. Both our work and PSGN focus on reconstruction on point clouds, where the point-set based metrics are straightforward and intuitive. 

\begin{figure}[!t]
\vspace{-5pt}
\centering
 \includegraphics[width=\linewidth]{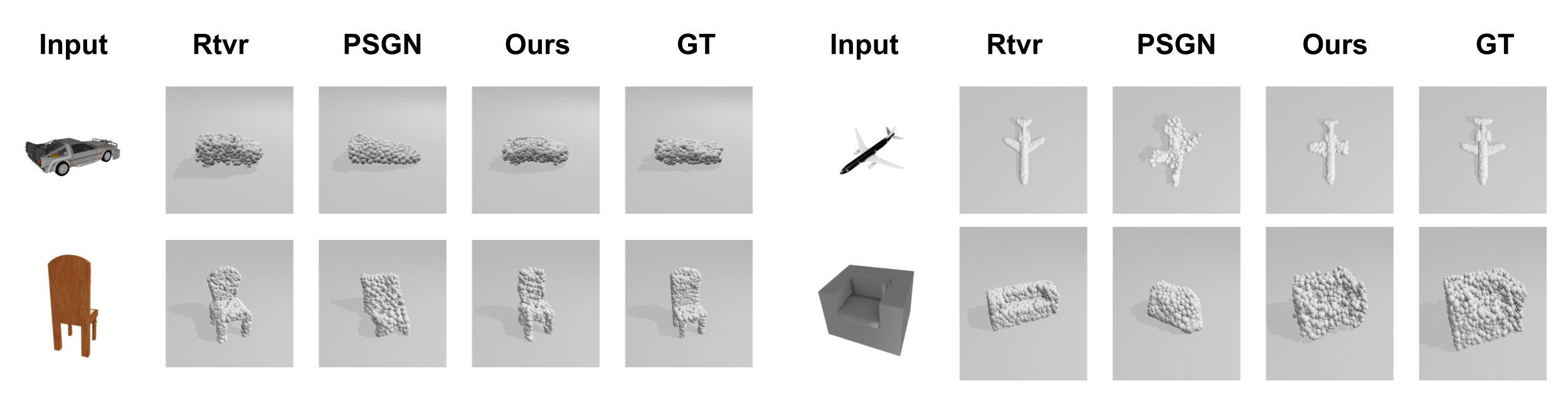}
 \caption{Visual comparison of different approaches. Left to right: input image, retrieved shape, point set generation network output, the output reconstruction trained with CD on our full model, ground truth shape. The examples are hand-picked from 4 categories.}
 \label{fig:examples}
 \vspace{-15pt}
\end{figure}

To contrast the two methods clearly, we show the visual comparison in Figure~\ref{fig:examples}. In general, our reconstruction can recover the main features from the object, even when the input template is not ideal for the image. This can be attributed to the combined benefits of shape retrieval and deformation - shape retrieval alone provides very good complete 3D shape templates without missing features, and deformation is able to preserve all of the template's main features while tweaking them to more closely match their shape in the image input.

\vspace{-5pt}
\subsection{Ablation Analysis}
\vspace{-5pt}
\noindent \textbf{Sensitivity to input shape template.}
Image-based shape retrieval could provide reasonable CAD model as template to start deforming with. We therefore also provide analysis on the degree to which \algoName relies on having a good shape template. 

\begin{wrapfigure}{r}{0.4\textwidth}
\vspace{-10pt}
\centering
 \includegraphics[width=\linewidth]{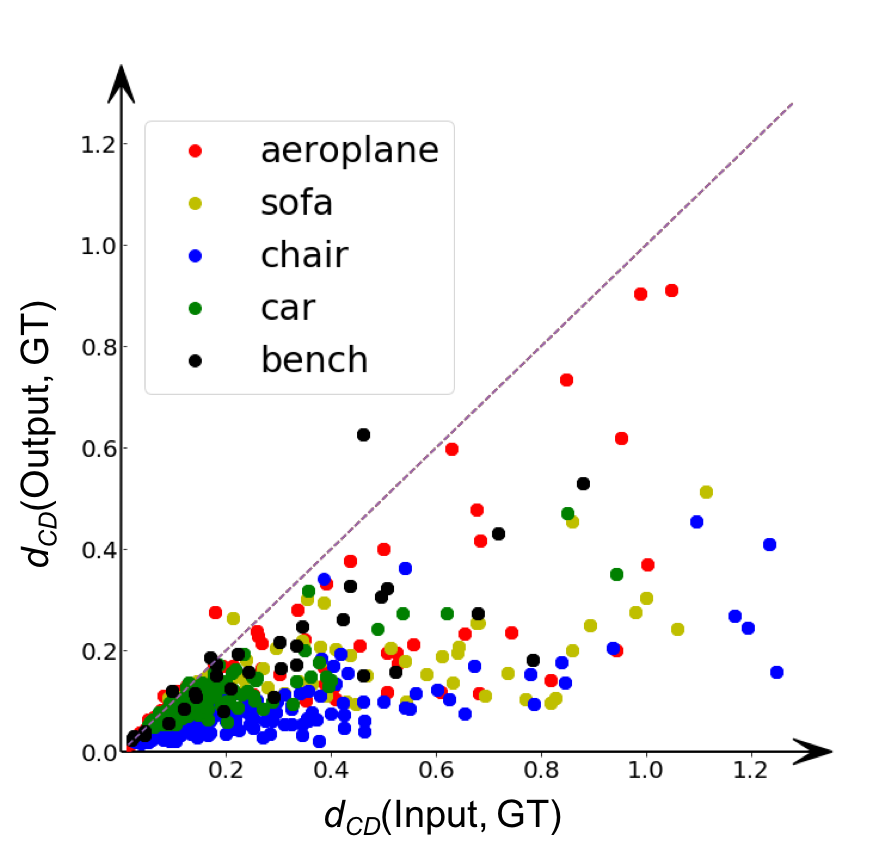}
 \vspace{-7pt}
 \caption{Sensitivity to shape template choice. The horizontal axis is the CD between ground truth and a random template input, and the vertical axis is the CD between ground truth and the output from {\algoName}.}
 \label{fig:sensitivity}
 \vspace{-20pt}
\end{wrapfigure}

To do this analysis, we compute the statistics on test set of all categories: for each group of input template (In), output (Out), ground truth shapes, we compute the tuple $(d_{CD}(\text{In}, \text{GT}), d_{CD}(\text{Out}, \text{GT}))$. Figure \ref{fig:sensitivity} shows the scattering plot of 516 random groups. Most of them lie closely to the horizontal axis, showing that \algoName is not very sensitive to the input template's quality, which indicates the robustness of \algoName. We also report the CD and EMD with randomly picked shape template as input (without shape retrieval) in \ref{table:ablation}, averaged on all categories. Note that {\algoName} without shape template retrieval still outperforms PSGN on average.

\noindent \textbf{Joint skip connections.} 
The joint skip connections stacking activations from 3D encoder and 3D decoder serve as information conveyors on the same level of spatial resolution. To verify its importance, we trained a network without the joint skip connections and measured its performance in CD and EMD metrics as shown in Table \ref{table:ablation}. Since the joint skip connections feed sharp information from the shape directly to deeper layers in the decoder, the reconstruction is more accurate in terms of CD loss.

\begin{wraptable}{r}{0.25\textwidth}
\centering
\vspace{-12pt}
\resizebox{\linewidth}{!}{
    \begin{tabular}{l|l|l}
    \hline
               & \cellcolor[HTML]{CBCEFB} CD   & \cellcolor[HTML]{FFC72C}EMD  \\ \hline
   \rowcolor[HTML]{E0E0E0} 
     w/o skip & 0.183 & 0.563 \\
     w/o reg    & 0.125 & 0.582 \\
     full    & 0.127 & 0.585 \\
     \rowcolor[HTML]{E0E0E0} 
    \end{tabular}
}
\caption{Ablation analysis}
\label{table:ablation}
\vspace{-10pt}
\end{wraptable}

\noindent \textbf{Regularizer.} To understand the functionality of the deformation regularizer, we have trained and tested a network without the regularization term in the objective function. Quantitatively, the model without regularization performs slightly better than the model with it, as shown in Table \ref{table:ablation}. However, the regularizer encourages more conservative and smooth deformation. For mesh reconstruction, this lowers the chance for faces to cross during deformation. As illustrated in the mesh reconstruction experiment in Figure \ref{fig:mesh}, the regularizer enforces the consistency among offsets of neighboring control points, which is not guaranteed in the model without the regularizer.

\begin{figure}[t]
\centering
 \includegraphics[width=\linewidth]{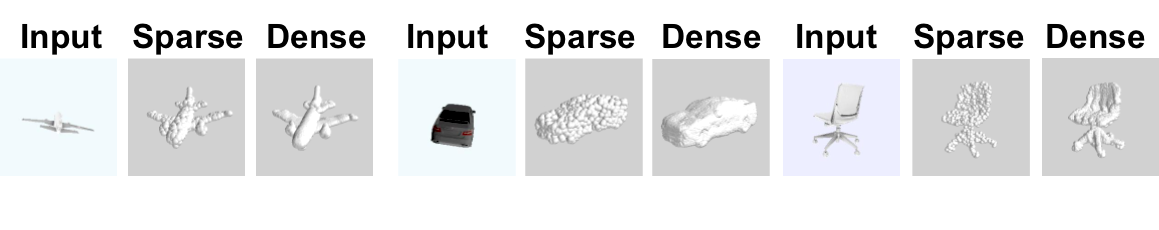}
 \vspace{-25pt}
 \caption{Reconstruction on sparse and dense point clouds. Sparse point cloud have 1024 points, dense point clouds have 16384 points. The model is trained on sparse point clouds only.}
 \label{fig:dense}
 \vspace{-15pt}
\end{figure}

\noindent \textbf{Point cloud density.} One advantage of FFD is that the same deformation can be applied to any number of points in the control points grid, which bridges the gap between low and high resolution. To have a fair comparison with PSGN, we also trained on sparse point clouds with 1024 points, but our trained model could be directly used on deforming and reconstructing dense point clouds. Figure \ref{fig:dense} shows the comparison between reconstructed sparse and dense point clouds, using the model trained on sparse point clouds. In general, network deforms dense point clouds similarly to sparse ones.

\vspace{-5pt}
\subsection{3D Reconstruction with Real Images}
\vspace{-15pt}
\begin{figure}[!h]
\vspace{-5pt}
\centering
 \includegraphics[width=\linewidth]{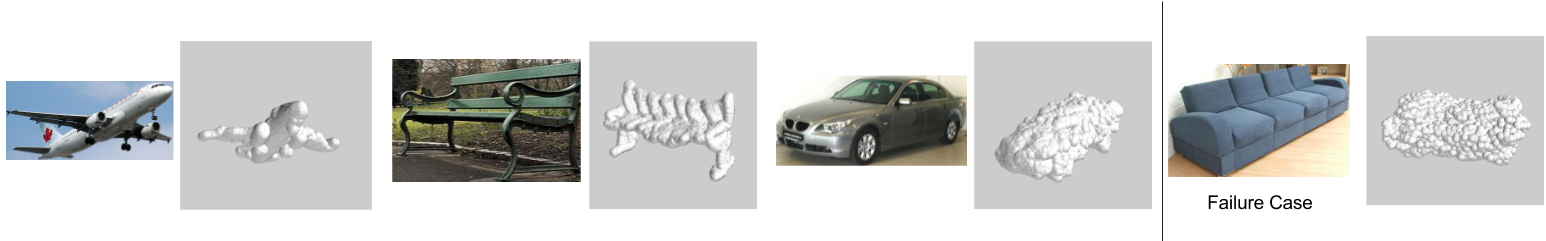}
 \vspace{-15pt}
 \caption{Real image reconstruction. On the first three test examples, {\algoName} generates reasonable reconstruction, while the last one fails.}
 \label{fig:real}
 \vspace{-10pt}
\end{figure}

We also tested our model on real world images and visualization results are shown in Figure \ref{fig:real}, including a failure case. Since we trained our model on synthesized images with single-colored backgrounds, we segmented real images as the input into network as done in PSGN. Our network successfully infers the 3D shape in some cases, but fails on some others. One solution could be domain adaptation and transfer learning, to bridge the gap between rendered and real world images, which we leave to future work.

\vspace{-5pt}
\subsection{Mesh Reconstruction}
\begin{figure}[ht]
\centering
\vspace{-10pt}
\begin{minipage}[c]{0.48\textwidth}
  \centering
    \includegraphics[width=\linewidth]{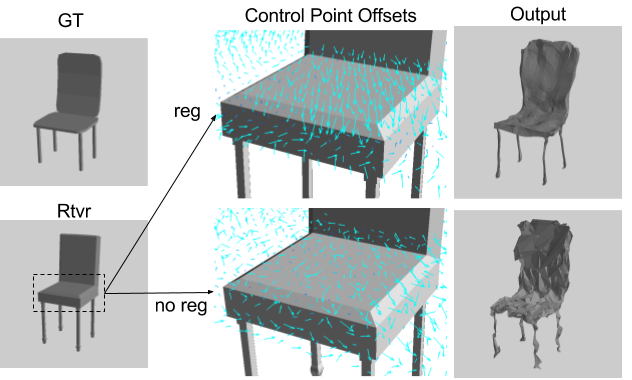}
  \end{minipage}\,
  \begin{minipage}[c]{0.48\textwidth}
  \centering
  \vspace{-5pt}
  \caption{Comparison between the behavior of DeformNet with and without regularization during training. The ground truth and retrieved model are similar except for the thickness of the seat and its height. The arrows shown in the zoomed-in figures are the offsets of control points. Color denote the offset's magnitude. The regularized model learns to consistently squeeze the seat uniformly whereas the unregularized model displaces the control points less smoothly, which results in the difference in the output mesh reconstruction.}
  \label{fig:mesh}
  \end{minipage}
  \vspace{-15pt}
\end{figure}

Free-form deformation has been widely used on deforming mesh objects, and with FFD layer in a deep neural network, the learnable manipulation and reconstruction on mesh become possible. As analyzed in section 4.3, the regularized model refrains from drastic changes in a local patch, which gives out mesh reconstruction with plausible quality. Figure \ref{fig:mesh} shows an example of mesh reconstruction on a chair.

\section{Conclusions and Outlook}
\vspace{-10pt}
This paper examines the structured prediction problem of regressing 3D point clouds based on image input to solve 3D Reconstruction with a single image. We note that existing methods with volumetric representations fall short on  computational efficiency. 
This paper leverages the Point cloud-based representation coupled with a 3D generalization of the Spatial Transformer using Free-Form Deformation to achieve state of the art results on reconstruction with ShapeNet.  
We introduce Free Form Deformation as a differentiable layer to enable 3D data manipulation. This can have wider implications beyond 3D reconstruction, such as in point-cloud processing, and learning to perform grasping on unseen objects.


\renewcommand*{\bibfont}{\small}
\printbibliography 

\newpage
\section{Supplementary Material}

\subsection{CD vs EMD}

\begin{wraptable}{r}{0.25\textwidth}
\vspace{-12pt}
\centering
\caption{Comparison of performance with CD and EMD. This evaluation is on a slightly different test set than in Table 1 and 2.}
\label{table:cross_eval}
\resizebox{\linewidth}{!}{
    \begin{tabular}{l|l|l}
    \cellcolor[HTML]{FFFFFF} Train \textbackslash \hspace{0.1em} Test & \cellcolor[HTML]{CBCEFB}CD & \cellcolor[HTML]{FFC72C}EMD \\ \hline
    \cellcolor[HTML]{CBCEFB}CD                        & 0.10   &  0.52   \\ 
    \rowcolor[HTML]{E0E0E0} 
    \cellcolor[HTML]{FFC72C} EMD                       & 0.11   &  0.50    \\ \hline
    \end{tabular}
}
\end{wraptable}
We trained DeformNet trained with both CD and EMD as objective functions, then test both of them using CD and EMD metrics. The average distance is reported on the Table 3. In terms of the point-set based metrics, the performance of these two models are similar. Besides, from our observation on output visualization, we don't find significant difference between these two metrics. With respect to computation efficiency, CD runs much faster with KDTree searching for nearest neighbors, so we primarily used it for the evaluations in this paper.

\subsection{Additional Images}
We present 18 additional qualitative examples, drawn at random from the test set:
\begin{figure}[!h]
\centering
\includegraphics[width=0.6\linewidth]{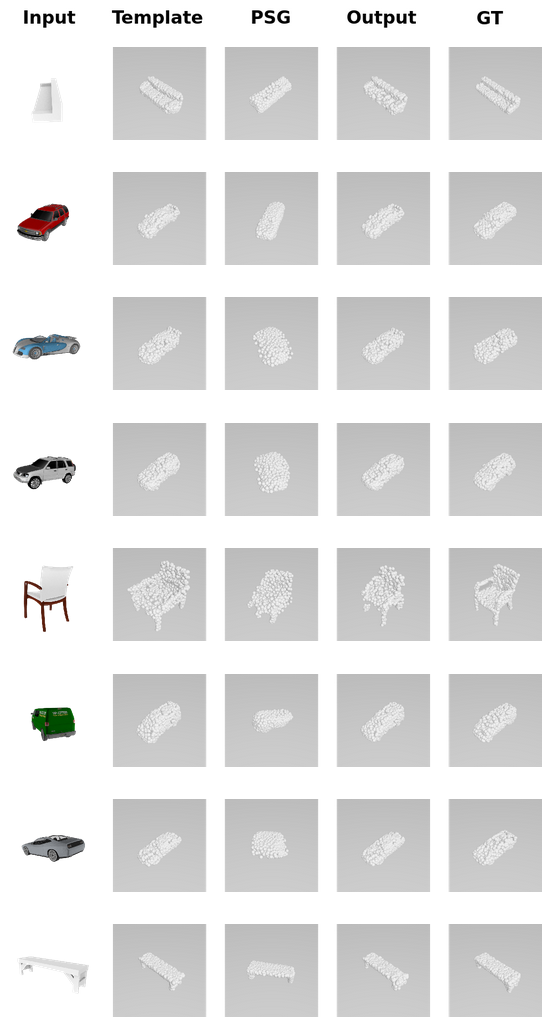}
\end{figure}

\begin{figure}[!h]
\centering
\includegraphics[width=0.66\linewidth]{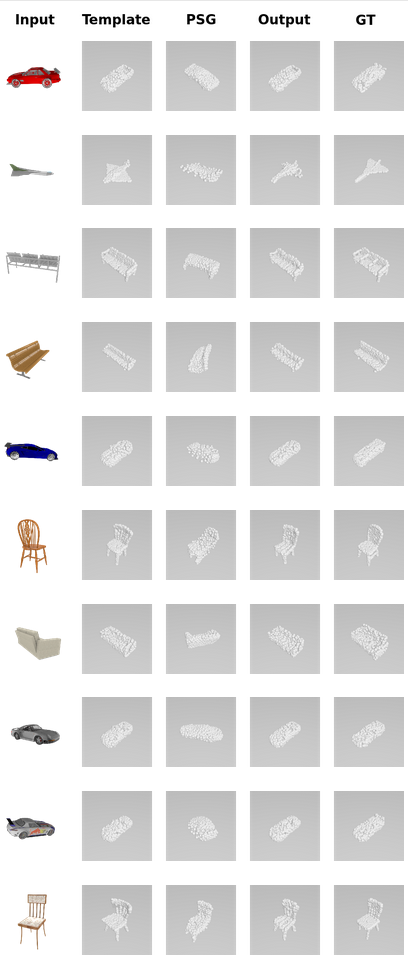}
\end{figure}

\end{document}